\documentclass[conference]{IEEEtran}
\IEEEoverridecommandlockouts
\usepackage{cite}
\usepackage{amsmath,amssymb,amsfonts}
\usepackage{algorithmic}
\usepackage{graphicx}
\usepackage{textcomp}
\usepackage{xcolor}
\usepackage{multirow}
\definecolor{Gray}{gray}{0.9}
 \usepackage{colortbl} 
 \usepackage{booktabs}

\def\BibTeX{{\rm B\kern-.05em{\sc i\kern-.025em b}\kern-.08em
    T\kern-.1667em\lower.7ex\hbox{E}\kern-.125emX}}
\begin{document}

\title{Personalization in Human Activity Recognition} 



\author{
\IEEEauthorblockN{Anna Ferrari, Daniela Micucci, Marco Mobilio, and Paolo Napoletano}
\IEEEauthorblockA{Department of Informatics, Systems and Communication, 
University of Milano Bicocca\\
Email: a.ferrari34@campus.unimib.it, \{daniela.micucci, marco.mobilio, paolo.napoletano\}@unimib.it}
}

\newcommand{\marco}[1]{\textcolor{orange}{{\it Marco: #1}}}

\maketitle

\begin{abstract}
In the recent years there has been a growing interest in techniques able to automatically recognize activities performed by people. This field is known as Human Activity recognition (HAR). HAR can be crucial in monitoring the wellbeing of the people, with special regard to the elder population and those people affected by degenerative conditions. One of the main challenges concerns the diversity of the population and how the same activities can be performed in different ways due to physical characteristics and life-style. In this paper we explore the possibility of exploiting physical characteristics and signal similarity to achieve better results with respect to deep learning classifiers that do not rely on this information.
\end{abstract}


\section{Introduction}
Nowadays smartphones are able to acquire, store, share, and elaborate huge amount of data in a very short time. As consequence of this technological development, new instruments related to data availability, data processing, and data analysis have attracted the attention of many research field including Human Activity Recognition (HAR). Indeed, the possibility to monitor people daily activities, risky activities, or changes in behavior (e.g. falls or diseases development) with a simple smartphone is very attractive and actual.

The increased computational power makes possible to consider not only traditional machine learning, but also more complex deep learning techniques.


Traditional machine learning methods (ML) are low cost in terms of time consumption, data, and complexity, however the dependency on expert knowledge in the features extraction phase often generates weak models difficult to compare \cite{zhu2019efficient,friday2018deep,yu2018multi}. On the other side, deep learning methods (DL) remain stable in terms of feature extraction, which is mainly automatically done, but the training phase requires more data, and, consequently, it is either very time consuming or requires expansive hardware \cite{ferrari2019-hand-crafted}. 

Regardless of the underlaying learning method (either classic machine learning or deep learning), real-world HAR systems achieve non satisfying recognition accuracy in real world applications mostly because HAR techniques struggle to generalize to new users and/or new environments \cite{hong2016toward,igual2015comparison}. One of the most relevant difficulty to face with new situations is due to the population diversity problem \cite{lane2011enabling}, that is, the natural differences between users when they perform the same activities. According to Zunino et al. \cite{zunino2017revisiting}, two factors are the reasons why the same activity is carried out in a different way.

\begin{itemize}
\item \emph{Inter-subject variability}, which refers to anthropometric differences of body parts or to incongruous personal styles in accomplishing the scheduled action. 
\item \emph{Intra-subject variability}, which represents the random nature of a single action class and reflects the fact that the same subject never performs an action in the same way.
\end{itemize}

To face subjects variability, algorithms should be trained on a representative number of subjects and on as many cases as possible. The number of subjects present in the dataset does not just impact the quality and robustness of the induced model, but also the ability to evaluate the consistency of results across subjects~\cite{lockhart2014limitations}. Nevertheless, in the sensor-based HAR community, datasets are in a low number. 

To overcome this issue, we propose machine learning and deep learning models based on user personalization, that is, based on the similarity of the subjects in terms of both physical characteristics and signals. In order to evaluate our approach, we compare our results with 
deep learning techniques. The impact of personalization in traditional ML has been covered in~\cite{ferrari2020personalization}.

\section{Proposed Methods}
To take into account the population diversity, we introduce the concept of similarity between subjects. The similarity between subjects is used to weight the training data in order to give more importance to data that are more similar to the data of the user under test.

Each subject $i$ can be described with a feature vector $\mathbf{g}_{i}=\{g_{1},\dots,g_{K}\}$. 
Similarity between two subjects $i$ and $j$ is defined as follows.
\vspace{-3mm}
\begin{center}
\begin{equation}\label{eq:1}
\text{sim}(i,j)  = e^{-\gamma d(i,j)}
\end{equation}
\end{center}

where $\gamma$ is a scale parameter and $d(i,j)$ is the Euclidean distance between the feature vectors of two subjects:

\begin{center}
\begin{equation}\label{eq:2}
\centering
d(i,j)=\sqrt{\sum_{k=1}^{K}(g_{k,i}-g_{k,j})^2}
\end{equation}
\end{center}

The resulting similarity value ranges from 0 to 1 where 0 means that the two subjects are dissimilar, and 1 means that the two subjects are equal.
The idea is to take advantage of the similarity between subjects in ML and DL engines as follows.

\begin{itemize}
\item \textbf{Personalized Machine Learning (PML)}. Given a subject $i$ under test, all the training data are weighted by using the similarity between the user $i$ and the rest of the users. 
We can define three types of similarity: \textit{physical-based} ($\text{sim}^{physical}$), \textit{sensor-based} ($\text{sim}^{sensor}$), and \textit{physical combined with sensor-based similarity} ($\text{sim}^{physical+sensor}$).
\item \textbf{Personalized Deep Learning (PDL)}. Starting from a minimum value $m$ we select the most $m$ similar subjects, with respect to the test subject. The network is trained with the samples related to these $m$ subjects. We selected the parameter $m$ equals to $10$, trained the network, and added 5 subjects until the maximum number of subjects in the dataset is achieved. 
\end{itemize}




\section{Experimental setup}
The PML technique in our experiment is an Adaboost classifier, while for the DL and the PDL we implemented two Convolutional Neural Networks.

PML, PDL and DL models have been trained and tested with two different splits, that is, Subject Independent (SI) and hybrid (HYB). In SI fashion the data of the user is left out from the training dataset, while in HYB fashion a part of the test user is insert in the training dataset.


Two public datasets containing accelerometer signals of Activities of Daily Living (ADLs) and Falls recorded with smartphones have been considered.

 \textbf{UniMiB-SHAR}~\cite{micucci2017unimib}  contains tri-axial acceleration data organized in 3s windows around the peak. The dataset contains 17 different activities (both ADLs and Falls) performed by 30 subjects. Sex, age, weight, and height of each subject are known. The original sampling rate is 50Hz. 
The subjects placed the smartphone used for the acquisition (a Samsung Galaxy Nexus I9250) half of the times in the left trouser pocket and the remaining times in the right one. 
 
%
 \textbf{Motion Sense}~\cite{malekzadeh2018protecting} contains time-series data generated by the accelerometers in an iPhone 6s worn by 24 participants. Each of the subjects performed 6 activities (only ADLs). The smartphone were kept in the participant's front pocket.
\\

\section{Results}

Table~\ref{tab:results1RQ4} shows preliminary results in terms of macro average accuracy (i.e., the average across subjects, split, and $m$ selection of subjects). Comparison between PML and PDL lead to contrasting conclusions. In UniMiB-SHAR, PML shows better results, while for Motion Sense is the opposite. Concerning DL without personalization, we see that in most of the cases, it outperforms both PML and PDL techniques. Results in UniMiB-SHAR remains contrasting.

\begin{table}[h]
\caption{Experimental Results -  accuracy of PDL and of PML.\label{tab:results1RQ4}}
\centering
\begin{tabular}{lccccc}
\toprule
&\multicolumn{2}{c}{\textbf{UniMiB-SHAR}}&\multicolumn{2}{c}{\textbf{Motion Sense}}\\
&PDL - PML & DL & PDL - PML & DL\\
\midrule
SI-physical &30.00 - 57.39 &\multirow{3}{*}{ \textbf{58.88}}&76.57 - 72.45& \multirow{3}{*}{\textbf{81.03}}\\
SI-sensor &42.08 - 57.00&&77.51 - 74.03\\
SI-physical sensor &42.09 - 56.93 &&77.51 - 73.85\\
HYB-physical & 44.42 - 85.44&\multirow{3}{*}{69.72}&79.06 - 77.76&\multirow{3}{*}{\textbf{85.75}}\\
HYB-sensor &46.62 - \textbf{84.71}  &&79.65 - 78.06\\
HYB-physical sensor &46.27 - \textbf{84.87}&&79.81 - 77.86\\
\bottomrule
\end{tabular}
\end{table}

\section{Conclusion}
Over last decades, HAR has been a very active field producing abundance of machine learning and deep learning based results. Nevertheless the lack of availability of large datasets prevent the traditional algorithms to generalize in real world situation. PML and DL techniques are becoming more and more popular because of their promising results. In this study we showed that traditional deep learning outperform personalized technique in most of the cases. Although, results on UniMiB-SHAR still confirm that personalized machine learning can yield better results. Given the contrasting results obtained with UniMiB-SHAR and Motion Sense datasets, we planned further investigation using other datasets, such as, for instance, MobiAct~\cite{vavoulas2016mobiact}.

\bibliographystyle{IEEEtran}
\bibliography{IEEEabrv,IEEErefe}

\end{document}